%% file: paper.tex
\def\year{2021}\relax
\documentclass[letterpaper]{article} 
\usepackage{aaai21}  
\usepackage{times}  
\usepackage{helvet} 
\usepackage{courier}  
\usepackage[hyphens]{url}  
\usepackage{graphicx} 

\usepackage{natbib}  
\usepackage{caption} 
\usepackage{amsfonts}
\usepackage{algorithm}
\usepackage{amsmath}
\usepackage{amssymb}
\usepackage{booktabs}
\usepackage{multirow}
\usepackage{multicol}

\usepackage{multirow}
\usepackage{epsfig,endnotes}
\usepackage{subfig}
\usepackage{grffile}
\usepackage[font=bf]{caption}
\usepackage{color, url}
\usepackage{xspace} 
\usepackage{mathrsfs}
\usepackage{amssymb}
\usepackage{amsmath}
\usepackage{amsthm}
\usepackage{epstopdf}
\usepackage{balance}
\usepackage{bm}
\usepackage{rotating}

\usepackage{enumitem}
\usepackage{makecell}

\setlist{nolistsep}

\newtheorem{theorem}{Theorem}

\newtheorem{definition}[theorem]{Definition}
\usepackage{algorithm}
\usepackage{algorithmic}

\def\UrlFont{\rm}  
\newcommand{\myparatight}[1]{\noindent{\bf {#1}:}~}

\usepackage[switch]{lineno}

\graphicspath{ {../fig/} }

\title{Semi-Supervised Node Classification on Graphs: Markov Random Fields \\ vs. Graph Neural Networks}

\author{
    Binghui Wang, Jinyuan Jia, Neil Zhenqiang Gong\\
}

\affiliations{
    Department of Electrical and Computer Engineering\\
    
    Duke University, Durham, NC 27708\\
    \{binghui.wang, jinyuan.jia, neil.gong\}@duke.edu

}

\begin{document}


\maketitle

\input{abstract}
\input{intro}
\input{background}

\input{method}

\input{exp}

\input{related}

\input{conclusion}

\bibliography{refs}

\input{appendix}

\end{document}

%% file: abstract.tex
\begin{abstract}
Semi-supervised node classification on graph-structured data has many applications such as fraud detection, fake account and review detection, user’s private attribute inference in social networks, and community detection.
Various methods such as pairwise Markov Random Fields (pMRF) and graph neural networks were developed for semi-supervised node classification.  
 pMRF is more efficient than graph neural networks. However, existing pMRF-based methods are less accurate than graph neural networks, due to a key limitation that they assume a heuristics-based constant edge potential for all edges. In this work, we aim to address the key limitation of existing pMRF-based methods. In particular, we propose to learn edge potentials for pMRF.  Our evaluation results on 
 various types of graph datasets
 show that our optimized pMRF-based method consistently outperforms existing graph neural networks in terms of both accuracy and efficiency. Our results highlight that previous work may have underestimated the power of pMRF for semi-supervised node classification.

\end{abstract}

%% file: intro.tex
\section{Introduction}

Given an undirected graph, a feature vector for each node in the graph, and a small set of labeled nodes, semi-supervised node classification aims to classify the remaining unlabeled nodes simultaneously. Semi-supervised node classification has many applications such as fraud detection~\cite{pandit2007netprobe,chau2011polonium,ye2011combining,tamersoy2014guilt,HooiFRAUDARACM}, fake account and review detection~\cite{wang2011review,sybilrank,akoglu2013opinion,rayana2015collective,wang2017sybilscar,wang2017gang,gao2018sybilfuse,wang2019structure},  user's private attribute inference in social networks~\cite{gong2014joint,jia2017attriinfer,gong2018attribute,wang2019graph}, and community detection~\cite{he2018network,jin2019graph,jin2019incorporating}. Therefore, many methods have been proposed for semi-supervised node classification. These  include conventional methods such as label propagation (LP)~\cite{zhu2003semi}, manifold regularization (ManiReg)~\cite{belkin2006manifold}, deep semi-supervised embedding (SemiEmb)~\cite{weston2012deep}, iterative classification algorithm (ICA)~\cite{lu2003link}, and pairwise Markov Random Fields (pMRF)~\cite{gatterbauer2017linearization}, as well as recent graph neural networks~\cite{scarselli2009graph,duvenaud2015convolutional,kipf2017semi,hamilton2017inductive,battaglia2018relational,xu2018representation,gao2018large,ying2018hierarchical,velivckovic2018graph,wu2019simplifying,qu2019gmnn,ma2019flexible,wang2020nodeaug}. 

In particular, pMRF associates a discrete random variable with each node to model its label. pMRF defines a \emph{node potential} for each node, which is a function of the node's random variable and models the node's label using the node's feature vector. pMRF also defines an \emph{edge potential} for each edge $(u, v)$, which is a function of the two random variables associated with nodes $u$ and $v$. Finally, pMRF defines a joint probability distribution of all random variables as the product of the node potentials and edge potentials. The edge potential for an edge $(u,v)$ is essentially a matrix, whose $(i, j)$th entry indicates the likelihood that $u$ and $v$ have labels $i$ and $j$, respectively. Therefore, an edge potential is also called \emph{coupling matrix}. Moreover, the standard Belief Propagation~\cite{Pearl88} or linearized Belief Propagation~\cite{gatterbauer2015linearized,gatterbauer2017linearization,jia2017attriinfer} were often used to perform inference on the pMRF. Compared to graph neural networks, pMRF inferred via linearized Belief Propagation is much more efficient in terms of running time. However, existing pMRF-based methods~\cite{pandit2007netprobe,tamersoy2014guilt,akoglu2013opinion,rayana2015collective,jia2017attriinfer,wang2017gang} are often less accurate than graph neural networks, due to a key limitation: they  assume a heuristics-based constant edge potential or coupling matrix for all edges.

In this work, we aim to address the limitation of existing pMRF-based methods. In particular, we propose a novel method to learn the coupling matrix. Specifically, we formulate learning the coupling matrix as an optimization problem, where the objective function is to learn a coupling matrix such that 1) the training loss is small (modeled as cross-entropy loss) and 2) the coupling matrix of an edge $(u,v)$ is consistent  with the predicted labels of nodes $u$ and $v$ (modeled as a regularization term). Moreover, we propose a novel algorithm to  solve the optimization problem. 
Our results on various types of graph datasets
show that  our method is more accurate and efficient than existing graph neural networks. 

%% file: background.tex
\section{Problem Definition and Background}

\subsection{Problem Definition}
We aim to solve 
the following \emph{semi-supervised node classification problem} on graph-structured data. 

\begin{definition}
Suppose we are given an undirected graph $G=(V, E)$ with $n=|V|$ nodes and $m=|E|$ edges. Each node $v$ has an $F$-dimension feature vector $\mathbf{x}_v$, and each node has a label from a label set  $\{1, 2, \cdots,  C\}$. 
 Moreover, we are given a small set of labeled nodes $L$, which we call training dataset. Our goal is to predict the labels of the remaining unlabeled nodes using the nodes' feature vectors and graph structure. 
\end{definition}
 
\subsection{Pairwise Markov Random Field (pMRF)} 
\label{backgroundpMRF}
pMRF was traditionally widely used for semi-supervised node classification in various applications such as auction fraud detection~\cite{pandit2007netprobe}, fake accounts and reviews detection~\cite{wang2011review,sybilrank,akoglu2013opinion,rayana2015collective,wang2017sybilscar,wang2017gang,wang2019graph}, malware detection~\cite{tamersoy2014guilt}, and 
user's private attribute inference
in social networks~\cite{gong2014joint,jia2017attriinfer,gong2018attribute,wang2019graph}. Specifically, we associate a random variable $r_v$ with each node $v$ to model its label, where $r_v=i$ means that $v$'s label is $i$. pMRF essentially models the joint probability distribution of the random variables $r_1, r_2, \cdots, r_n$, where  the statistical correlations/independence between the random variables are captured by the graph structure. Formally, pMRF models the joint probability distribution as follows:
\begin{align}
\label{pMRF}
\text{Pr}(r_1, r_2, \cdots, r_n) \propto \prod_{v\in V}\phi_v(r_v) \prod_{(u,v)\in E} \psi_{uv}(r_u,r_v),
\end{align}
where $\phi_v(r_v)$ is called node potential and $\psi_{uv}(r_u,r_v)$ is called edge potential. For convenience, we represent the node potential $\phi_v(r_v)$ as a probability distribution $\tilde{\mathbf{{q}}}_v$ over the $C$ labels $\{1, 2, \cdots,  C\}$. Specifically,  $\tilde{\mathbf{{q}}}_v$ is a row vector and $\tilde{q}_v(i)=\phi_v(r_v=i)$ for $i\in\{1, 2, \cdots,  C\}$. 
If nodes' feature vectors are available, $\tilde{\mathbf{{q}}}_v$ can be learnt.
In particular, given the labeled nodes $L$ and nodes' feature vectors, we learn a multi-class logistic regression classifier and use the classifier to predict the probability distribution  $\tilde{\mathbf{{q}}}_v$ for each node. 
If nodes's feature vectors are unavailable, we can assign $\tilde{\mathbf{{q}}}_v$ based on the labeled nodes $L$. 
Specifically, if node $v$ is a labeled node, $\tilde{\mathbf{{q}}}_v$ is set as the node $v$'s one-hot label vector; otherwise, all entries in $\tilde{\mathbf{{q}}}_v$ have the equal value $1/C$. 
Moreover, we represent the edge potential $\psi_{uv}(r_u,r_v)$ as a \emph{symmetric} matrix $\tilde{\mathbf{{H}}}_{uv}\in \mathbb{R}^{C\times C}$, where $\tilde{{{H}}}_{uv}(i,j)=\psi_{uv}(r_u=i,r_v=j)$ for $i, j\in\{1, 2, \cdots,  C\}$. $\tilde{\mathbf{{H}}}_{uv}$ models the coupling strength between the labels of nodes $u$ and $v$, and thus $\tilde{\mathbf{{H}}}_{uv}$ is also called \emph{coupling matrix}. In particular, a larger $\tilde{{{H}}}_{uv}(i,j)$ means that $u$ and $v$ are more likely to have labels $i$ and $j$, respectively.

\myparatight{Inference via linearized Belief Propagation (LinBP)} Given the joint probability distribution in Equation~\ref{pMRF}, the 
probability distribution $\tilde{\mathbf{{q}}}_v$ for each node, and the coupling matrix $\tilde{\mathbf{{H}}}_{uv}$  for each edge, we infer the marginal probability distribution $\tilde{\mathbf{{p}}}_v$ (a row vector) for each node $v$ and use it to predict $v$'s label.  
Specifically, Belief Propagation (BP)~\cite{Pearl88} is a standard method to infer the marginal probability distributions. However, BP is not guaranteed to converge on loopy graphs and is not scalable due to message maintenance on each edge. Recent work~\cite{gatterbauer2015linearized,gatterbauer2017linearization} proposed to linearize BP (LinBP) to address these limitations. Given LinBP, the \emph{approximate} marginal probability distributions ${\mathbf{{p}}}_v$ are solutions to the following linear system~\cite{gatterbauer2017linearization}:
\begin{small}
\begin{align}
\label{LinBP}
{\mathbf{{p}}}_v &= {\mathbf{{q}}}_v + \sum_{u\in \Gamma_v} {\mathbf{{p}}}_u {\mathbf{{H}}}_{uv},\ \forall v\in V,
\end{align}
\end{small}%
where $\Gamma_v$ is the set of neighbors of $v$,
 ${\mathbf{{q}}}_v =\tilde{\mathbf{{q}}}_v-\frac{1}{C}$, and  ${\mathbf{{H}}}_{uv} =\tilde{\mathbf{{H}}}_{uv}-\frac{1}{C}$. Roughly speaking,  ${\mathbf{{q}}}_v$ and ${\mathbf{{H}}}_{uv}$ are centered versions of  $\tilde{\mathbf{{q}}}_v$ and $\tilde{\mathbf{{H}}}_{uv}$, respectively. Note that ${\mathbf{{p}}}_v$ solved by LinBP from Equation~\ref{LinBP} is not necessarily a probability distribution any more, e.g., it may not sum to 1. The label of $v$ is predicted as the one that has the largest value in  ${\mathbf{{p}}}_v$.

\myparatight{Limitation} The key limitation of existing pMRF-based semi-supervised node classification methods is that they assume a heuristics-based constant coupling matrix for all edges. For example, $\tilde{\mathbf{{H}}}_{uv}(i,i)=0.9$ and $\tilde{\mathbf{{H}}}_{uv}(i,j)=\frac{0.1}{C-1}$, where $i,j\in \{1, 2, \cdots,  C\}$ and $i\neq j$, which means that two linked nodes are more likely to have the same label and does not distinguish between labels and between edges. In this work, we address this limitation via proposing a method to learn the coupling matrix. Moreover, we empirically show that, after learning the coupling matrix, pMRF outperforms state-of-the-art graph neural networks on multiple benchmark datasets.

%% file: method.tex
\section{Learning the Coupling Matrix (LCM)}

\subsection{Formulating an Optimization Problem}

\subsubsection{Constraining the coupling matrices:}
In pMRF, each edge has a coupling matrix $\mathbf{H}_{uv}$. Therefore, it is challenging to learn a different coupling matrix for each edge as the number of parameters is much larger than the training dataset size. To address the challenge, we constrain the coupling matrices of different edges as different scalings of the same coupling matrix. In particular, we assume $\mathbf{H}_{uv}=W_{uv}\mathbf{H}$ for each edge $(u,v)\in E$. We model the scaling parameter $W_{uv}$ as the weight of the edge $(u,v)$ and we denote all the scaling parameters as a symmetric weight matrix ${\mathbf{W}}$ of the graph. Given such assumption on the coupling matrices, we can transform Equation~\ref{LinBP} as follows:
{
\footnotesize
\begin{align}
\label{LinBP-alter}
\mathbf{P} &= {\mathbf{{Q}}} +  \mathbf{W} \mathbf{P} \mathbf{{H}},
\end{align}
}%
where the matrix $\mathbf{P}=[\mathbf{p}_1; \mathbf{p}_2;\cdots;\mathbf{p}_n]$ and the matrix $\mathbf{Q}=[\mathbf{q}_1; \mathbf{q}_2;\cdots;\mathbf{q}_n]$. 
{According to~\cite{gatterbauer2015linearized}, Equation~\ref{LinBP-alter} has a solution when $\mathbf{I} - \mathbf{H} \otimes \mathbf{W}$ is invertible, where $\mathbf{I}$ is an identity matrix and $\otimes$ is the Kronecker product. 
LinBP uses the power iteration method to solve Equation~\ref{LinBP-alter} and it can find the solution when $\rho(\mathbf{H}) < \frac{1}{\rho(\mathbf{W})}$, where $\rho(\cdot)$ is the spectral radius of a matrix.\footnote{The spectral radius of a square matrix is the maximum of the
absolute values of its eigenvalues.}}

\subsubsection{Formulating an optimization problem:} 
Intuitively, we aim to achieve two goals when learning the coupling matrix and weight matrix. The first goal is that the labels predicted for the training nodes based on the leant  coupling matrix and weight matrix should match the ground truth labels of the training nodes. The second goal is that   the predicted label distributions $\sigma({\mathbf{p}}_u)$ and $\sigma({\mathbf{p}}_v)$ of nodes $u$ and $v$ should be \emph{consistent} with the coupling matrix $W_{uv}{\mathbf{H}}$ for each edge $(u,v)$. Specifically, if $u$ and $v$ are more likely to be predicted to have labels $i$ and $j$ (i.e., the probabilities $\sigma({\mathbf{p}}_u)_i$ and $\sigma({\mathbf{p}}_v)_j$ are larger), respectively, then the entry  $W_{uv}H(i,j)$ should be larger. This is because $W_{uv}H(i,j)$ 
encodes the likelihood that $u$ and $v$ have labels $i$ and $j$, respectively. 

To quantify the first goal, we adopt a loss function over the training dataset. Specifically, we consider a standard cross-entropy loss function $\textrm{L}({\mathbf{W}}, \mathbf{H})$ as follows:
{\footnotesize
\begin{align} 
\label{loss}
 \textrm{L}({\mathbf{W}}, \mathbf{H}) = - \sum_{l \in L} \mathbf{y}_l^T \log \sigma({{\mathbf{p}}_l}),
\end{align}
}%
where $\sigma$ is the softmax function, which is defined as $\sigma(\mathbf{p}_l)_i = \frac{\exp(\mathbf{p}_l(i))}{\sum_{j=1}^C \exp(\mathbf{p}_l(j))}$ and converts the vector $\mathbf{p}_l$ to be  a probability distribution;   
the vector $\mathbf{y}_l$ is the one-hot encoding of the node $l$'s label, i.e.,  if node $l$ belongs to the $i$th class, then ${y}_l(i)=1$ and ${y}_l(j)=0$, for all $j \neq i$. 
To quantify the second goal, we define the following term $\textrm{R}({\mathbf{W}}, \mathbf{H})$:
{\footnotesize
\begin{align}
\textrm{R}({\mathbf{W}}, \mathbf{H})  = - \sum_{(u,v)\in E}  \sigma({\mathbf{p}}_u) \cdot W_{uv}{\mathbf{H}} \cdot \sigma({\mathbf{p}}_v)^T.  
\end{align}
}%
$\textrm{R}({\mathbf{W}}, \mathbf{H})$ is smaller if the second goal is better satisfied, i.e., the predicted label distributions of two nodes are more consistent with the coupling matrix of the corresponding edge. Combining the two goals, we propose to learn the coupling matrix and weight matrix via solving the following optimization problem:
{\footnotesize
\begin{align} 
\label{objfunc}
\min_{{\mathbf{W}}, \mathbf{H}} \ \mathcal{L}({\mathbf{W}},\mathbf{H}) = \textrm{L}({\mathbf{W}}, \mathbf{H}) + \lambda \textrm{R}({\mathbf{W}}, \mathbf{H}),
\end{align}
}%
where $\lambda>0$ is a hyperparameter that balances between the two goals. 
Our term $\textrm{R}({\mathbf{W}}, \mathbf{H})$ can also be viewed as a regularization term, and we call it \emph{consistency regularization}. We note that a similar consistency regularization term was also proposed by \cite{wang2019graph}, but only for binary classification.

\subsubsection{Challenge for solving the optimization problem:} 
It is computationally challenging to solve the optimization problem in Equation~\ref{objfunc}. 
The reason is each $\mathbf{p}_v$ depends on every edge weight and each entry of the coupling matrix. For example, suppose we use gradient descent to solve the optimization problem. In each iteration, we need to compute the gradient $\frac{\partial \mathcal{L}({\mathbf{W}}, \mathbf{H})}{\partial {W_{uv}}}$, which involves the gradient $\frac{\partial {\mathbf{P}}}{\partial {W_{uv}}}$. 
However, according to Equation~\ref{LinBP-alter}, $\frac{\partial {\mathbf{P}}}{\partial {W_{uv}}}$ is a solution to the following system:
{\footnotesize
\begin{align}
\label{derivative}
\frac{\partial {\mathbf{P}}}{\partial {W_{uv}}} = {\mathbf{W}} \frac{\partial {\mathbf{P}}}{\partial {W_{uv}}} {\mathbf{H}} + \frac{\partial {\mathbf{W}}}{\partial {W_{uv}}} {\mathbf{P}} {\mathbf{H}}. 
\end{align}
}%

Therefore, in each iteration of gradient descent, we need to  solve the above system of equations for \emph{each} edge, which is computationally infeasible for large graphs.

\subsection{Solving the Optimization Problem}

We propose an approximate algorithm to efficiently solve the optimization problem. Our key observation is that $\mathbf{P}$ is often iteratively solved from Equation~\ref{LinBP-alter} as follows:
\begin{small}
\begin{align}
\label{LinBP-iter}
\mathbf{P}^{(t)} &= {\mathbf{{Q}}} +  \mathbf{W} \mathbf{P}^{(t-1)} \mathbf{{H}},
\end{align}
\end{small}%
where $\mathbf{P}^{(t)}$ is the matrix $\mathbf{P}$ in the $t$th iteration. Based on the observation, we propose to \emph{alternately} update  $\mathbf{P}$ and learn the weight matrix $\mathbf{W}$ and coupling matrix $\mathbf{H}$. Specifically, given the current $\mathbf{W}$, $\mathbf{H}$, and $\mathbf{P}$, we update $\mathbf{P}$ in the next iteration. Then, given the current  $\mathbf{P}$, we learn $\mathbf{W}$ and $\mathbf{H}$ such that 1) the matrix $\mathbf{P}$ in the next iteration has a small cross-entropy loss on the training dataset and 2) the consistency regularization term is large for the current $\mathbf{P}$.

\myparatight{Updating ${\mathbf{P}}^{(t)}$} 
Given the weight matrix ${\mathbf{W}}^{(t-1)}$, the coupling matrix $\mathbf{H}^{(t-1)}$, and the  matrix  ${\mathbf{P}}^{(t-1)}$ in the $(t-1)$th iteration, we  compute   ${\mathbf{P}}^{(t)}$ in the $t$th iteration as follows:
{
\footnotesize
\begin{align} 
\label{propgatestep}
{\mathbf{P}}^{(t)} = {\mathbf{Q}} + {\mathbf{W}}^{(t-1)} {\mathbf{P}}^{(t-1)} {\mathbf{H}}^{(t-1)}. 
\end{align}
}

\vspace{-2mm}
\myparatight{Learning ${\mathbf{W}}^{(t)}$ and $\mathbf{H}^{(t)}$} 
Given the weight matrix ${\mathbf{W}}^{(t-1)}$ and the coupling matrix $\mathbf{H}^{(t-1)}$ in the $(t-1)$th iteration as well as the matrix ${\mathbf{P}}^{(t)}$ in the $t$th iteration, we learn the weight matrix ${\mathbf{W}}^{(t)}$ and the coupling matrix $\mathbf{H}^{(t)}$  in the $t$th iteration as a solution to the following optimization problem: 
{\footnotesize
\begin{align} 
\label{weightlearning}
& \min_{\mathbf{W}^{(t)}, \mathbf{H}^{(t)}} \mathcal{L}({\mathbf{W}^{(t)}}, \mathbf{H}^{(t)}) = -\sum_{l \in L} \mathbf{y}_l^T \log \sigma({\mathbf{p}}_l^{(t+1)}) \nonumber \\ 
& \qquad \qquad - \lambda \sum_{(u,v)\in E} \sigma({\mathbf{p}}_u^{(t)}) \cdot W_{uv}^{(t)}{\mathbf{H}}^{(t)} \cdot (\sigma({\mathbf{p}}_v^{(t)}))^T, 
\end{align}
}%
where ${\mathbf{P}}^{(t+1)} = {\mathbf{Q}} + \mathbf{W}^{(t)} {\mathbf{P}}^{(t)} {\mathbf{H}}^{(t)}$. We leverage gradient descent to learn $\mathbf{W}^{(t)}$ and $\mathbf{H}^{(t)}$. Specifically, we have the following gradients:
{\footnotesize
\begin{align}
\frac{\partial \mathcal{L}({\mathbf{W}^{(t)}, \mathbf{H}^{(t)}})}{\partial {W_{uv}^{(t)}}} & = -\sum_{l \in L} \sum_{j=1}^C ({y}_l(j) - \sigma(\mathbf{p}_l^{(t+1)})_j) \frac{\partial {p}_l^{(t+1)}(j)}{\partial {W_{uv}^{(t)}}} \nonumber \\
& \quad - \lambda \sigma({\mathbf{p}}_u^{(t)}) \cdot {\mathbf{H}}^{(t)} \cdot (\sigma({\mathbf{p}}_v^{(t)}))^T,
\end{align}
}%
{\footnotesize
\begin{align}
& \frac{\partial \mathcal{L}(\mathbf{W}^{(t)}, {\mathbf{H}^{(t)}})}{\partial {H_{ij}^{(t)}}} = -\sum_{l \in L} \Big( ({y}_l(j) - \sigma(\mathbf{p}_l^{(t+1)})_j) \frac{\partial {p}_l^{(t+1)}(j)}{\partial {H_{ij}^{(t)}}} \nonumber \\ 
& \qquad + ({y}_l(i) - \sigma(\mathbf{p}_l^{(t+1)})_i) \frac{\partial {p}_l^{(t+1)}(i)}{\partial {H_{ij}^{(t)}}} \Big) \nonumber \\
& \qquad - \lambda \sum_{(u,v)\in E}  W_{uv}^{(t)} \cdot \sigma({\mathbf{p}}_u^{(t)})_i \cdot \sigma({\mathbf{p}}_v^{(t)})_j, 
\end{align}
}%
where we have:
{
\footnotesize
\begin{align}
\frac{\partial {{p}}_l^{(t+1)}(j)}{\partial {W_{uv}^{(t)}}} &=
\begin{cases}
 ({\mathbf{P}}^{(t)} {\mathbf{H}}^{(t)})_{vj}, &\ \text{ if } u= l \\
 ({\mathbf{P}}^{(t)} {\mathbf{H}}^{(t)})_{uj}, &\ \text{ if } v= l \\
 0, &\ \text{ otherwise,}
\end{cases}\\
\frac{\partial {p}_l^{(t+1)}(j)}{\partial {H_{ij}^{(t)}}} &= (\mathbf{W}^{(t)} \mathbf{P}^{(t)})_{li},\\
\frac{\partial {p}_l^{(t+1)}(i)}{\partial {H_{ij}^{(t)}}} &= (\mathbf{W}^{(t)} \mathbf{P}^{(t)})_{lj}.
\end{align}
}%
We have both ${\partial {p}_l^{(t+1)}(j)}/{\partial {H_{ij}^{(t)}}}$
and ${\partial {p}_l^{(t+1)}(i)}/{\partial {H_{ij}^{(t)}}}$
when computing ${\partial \mathcal{L}(\mathbf{W}^{(t)}, {\mathbf{H}^{(t)}})}/{\partial {H_{ij}^{(t)}}}$ 
because $\mathbf{H}$ is symmetric. We initialize $\mathbf{W}^{(t)}$ and $\mathbf{H}^{(t)}$ as $\mathbf{W}^{(t-1)}$ and $\mathbf{H}^{(t-1)}$, respectively. Then, we use gradient descent to update $\mathbf{W}^{(t)}$ and $\mathbf{H}^{(t)}$ with learning rates $\gamma_1$ and $\gamma_2$, respectively. 

{
We note that the learnt weight matrix $\mathbf{W}$ and coupling matrix $\mathbf{H}$ are bounded during optimization. 
This is because the edge potential matrix $\tilde{\mathbf{H}}_{uv}$ is non-negative and each row sums to 1. Moreover, $W_{uv} \mathbf{H} = \mathbf{H}_{uv}= \tilde{\mathbf{H}}_{uv} - 1/C$. Therefore, $W_{uv} \mathbf{H}$ is bounded. During optimization, we can clip $W_{uv}\mathbf{H}$ to satisfy such constraints if they are too large. However, we found that $W_{uv}\mathbf{H}$ is small even if we do not clip in our experiments. This is because of our initializations of $W_{uv}$ and $\tilde{\mathbf{H}}_{uv}$ 
and that our LCM method requires a small number of iterations. 
}

%% file: exp.tex
\section{Evaluation}

\subsection{Benchmark Datasets}

Following  previous work~\cite{yang2016revisiting,kipf2017semi}, we use three benchmark citation graphs (i.e., Cora, Citeseer, and Pubmed)~\cite{sen2008collective} and one benchmark knowledge graph NELL~\cite{carlson2010toward,yang2016revisiting}.
{We also adopt two real-world large-scale social graphs (i.e., Google+~\cite{Gong12-imc,jia2017attriinfer} and Twitter~\cite{wang2017sybilscar}) to evaluate our method. 
Descriptions of these datasets and their basic statistics are shown in our supplementary material.
}

\subsection{Training, Validation, and Testing} 
We split each dataset into a training dataset, a validation set, and a testing dataset. 
We use the validation dataset to tune hyperparameters and use the testing dataset for evaluation. 

For the citation graphs, previous work~\cite{kipf2017semi,velivckovic2018graph} has set 1,000 nodes in each graph as the testing dataset. For consistent comparisons, we adopt these testing datasets and fix them. Moreover, we sample 20 nodes from each class uniformly at random as the training dataset, and we sample 500 nodes in total uniformly at random as the validation dataset. For the NELL graph, previous work~\cite{yang2016revisiting} has set 969 nodes as the testing dataset, which we adopt. Moreover, like previous work~\cite{kipf2017semi}, we consider the extreme case where just one node from each class is sampled for training. 
We also randomly sample 500 nodes in total 
as the validation dataset. We repeat sampling the training and validation datasets 5 trials and report the average accuracies and standard deviations on the testing dataset.\footnote{Note that the results of the baseline methods reported in previous work only use one training and validation dataset without reporting the averages among multiple trials. Therefore, their results are different from what we show in Table~\ref{performance}.} 

{
For the Google+ dataset, for each city, we randomly select 1,000 positive users, i.e., who live/lived in the city, and 1,000 negative users, i.e., who do/did not live in the city, as the training dataset; we randomly select 1,000 positive users and 1,000 negative users as the validation dataset; and we treat the remaining users who disclosed at least one city as the testing dataset. 
Note that we have 50 cities in the Google+ dataset, where each city is treated as a binary classification problem.
The accuracy  on the Google+ dataset is averaged over the 50 cities.
For the Twitter dataset, we randomly sample 1,000 genuine  users and 1,000 fake users 
as the training dataset; we randomly sample 1,000 genuine users and 1,000 fake users 
as the validation dataset; 
and we randomly select another 5,000 genuine users and 5,000 fake users as the testing dataset. 
We sample a relatively larger number of training users in Google+ and Twitter, due to their large graph sizes.
}

\begin{table*}[t]
\centering
\addtolength{\tabcolsep}{-2pt}
\begin{tabular}{|c|c|c|c|c|c|c|c|}
\hline
\multicolumn{2}{|c|}{\bf \textbf{Methods}} & {\bf \textbf{Cora}}  &  {\bf \textbf{Citeseer}} &   {\bf \textbf{Pubmed}} & {\bf \textbf{NELL}} &   {\bf \textbf{Google+}} & {\bf \textbf{Twitter}} \\ \hline
\multirow{2}{*}{\bf \makecell{Graph \\ Embedding}} 
& {\bf DeepWalk} &  {0.669$\pm$0.017}  &  {0.428$\pm$0.019}  &   {0.651$\pm$0.015} & {0.352$\pm$0.013} & {--} & {--} \\  \cline{2-8}
& {\bf Planetoid} &  {0.746$\pm$0.011}  &  {0.645$\pm$0.029}  &   {0.699$\pm$0.035} & {0.435$\pm$0.015} & {--} & {--} \\  \hline \hline

\multirow{4}{*}{\bf \makecell{Graph \\ Neural \\ Network}} 
{\bf } & {\bf Chebyshev} &  {0.786$\pm$ 0.018}  &  {0.694$\pm$ 0.023} &   {0.701$\pm$ 0.024} & {--}  & {--} & {--} \\ \cline{2-8}
{\bf } & {\bf GCN} & {0.823$\pm$ 0.009} &   {0.714$\pm$ 0.016} &   {{0.756$\pm$ 0.032}} & {0.596$\pm$0.026} & {--} & {--} \\ \cline{2-8}
{\bf } & {\bf GraphSAGE} & {0.811$\pm$ 0.009} &   {0.664$\pm$ 0.010}  &   {0.757$\pm$  0.014} & {0.574$\pm$0.015} & {--} & {--} \\ \cline{2-8}
{\bf } & {\bf GAT} & {{0.824$\pm$ 0.021}} &   {0.708$\pm$ 0.016}  &  {0.755$\pm$ 0.030} & {0.578$\pm$0.010} & {--} & {--} \\ \hline  \hline

\multirow{2}{*}{\bf \makecell{Belief \\ Propagation}} 
{\bf } & {\bf BP} &  {{0.804$\pm$ 0.023}}  &  { {0.704$\pm$ 0.011}}  &   { {0.734$\pm$ 0.024}}  & {0.587$\pm$0.015} & {--} & {--} \\ \cline{2-8}
{\bf } & {\bf LinBP} &  { {0.809$\pm$ 0.014}}  &  { {0.707$\pm$ 0.004}}  &   { {0.744$\pm$ 0.028}}  & {0.588$\pm$0.013}  & {0.741$\pm$0.013} & {0.687$\pm$0.002} \\ \hline\hline
\multirow{4}{*}{\bf \makecell{Our \\ Methods}} 
{\bf } & {\bf LCM-wo} &  { {0.813$\pm$ 0.007}}  &  { {0.716$\pm$ 0.006}}  &   { {0.754$\pm$ 0.018}} & { {0.626$\pm$0.013}} & {0.757$\pm$0.012} & {0.712$\pm$0.002} \\ \cline{2-8}
{\bf } & {\bf LCM-L1} &  { {0.808$\pm$ 0.009}}  &  { {0.711$\pm$ 0.005}}  &   { {0.752$\pm$ 0.013}} & {0.609$\pm$0.008} & {0.749$\pm$0.018} & {0.699$\pm$0.004} \\ \cline{2-8}
{\bf } & {\bf LCM-L2} &  {{0.811$\pm$ 0.013}}  &  { {0.713$\pm$ 0.007}}  &   {{0.753$\pm$ 0.016}} & { {0.616$\pm$0.002}} & {0.751$\pm$0.016} & {0.709$\pm$0.004} \\  \cline{2-8}
{\bf } & {\bf LCM} &  {{0.833$\pm$ 0.007}}  &  {{0.722$\pm$ 0.005}}  &   {{0.770$\pm$ 0.019}} & {{0.647$\pm$0.005}} & {0.779$\pm$0.012} & {0.748$\pm$0.002} \\ \hline
\end{tabular} \\
\caption{Average accuracies and standard deviations of compared methods on the six graphs. 
''--'' means the methods cannot be executed on our machine as they 
require more memory than we have. {We note that all the compared graph embedding methods and graph neural networks cannot be executed on our machine for the two large-scale social graphs.}
} 
\label{performance}
\end{table*}

\begin{table*}[t]
\centering
\addtolength{\tabcolsep}{-2pt}
\begin{tabular}{|c|c|c|c|c|c|c|c|c|c|c|c|c|}
\hline
\multirow{2}{*}{\bf \bf{Dataset}} & \multicolumn{2}{c|}{\bf \bf Cora} & \multicolumn{2}{c|}{\bf \bf Citeseer} & \multicolumn{2}{c|}{\bf \bf Pubmed} & \multicolumn{2}{c|}{\bf \bf NELL} & \multicolumn{2}{c|}{\bf \bf Google+} & \multicolumn{2}{c|}{\bf \bf Twitter} \\ \cline{2-13}
			& homo & heter & homo & heter & homo & heter & homo & heter & homo & heter & homo & heter \\ \hline
\bf{\bf Average initialized weights}  &  0.225 &    0.197  &   0.315  &  0.347  &    0.138       &  0.147   &    0.039  &  0.035  & 0.107 & 0.091 & 0.022 & 0.017      \\ \hline
 \bf{\bf Average learnt weights} &    0.354   &    0.261   &   0.493   &   0.448 &    0.652       &  0.548   &   0.105  &  0.089   & 0.569 & 0.242 & 0.595 & 0.329   \\ \hline
\end{tabular}
\caption{Average initialized weights and weights learnt by our method  for homogeneous edges vs. heterogeneous edges on the six graphs. ``homo" and ``heter"  mean  homogeneous edges and heterogeneous edges, respectively. Our method learns larger average weights for homogeneous edges than heterogeneous edges. }
\label{ave_weight}
\end{table*}

\subsection{Compared Methods}
\label{comp_method}
We compare our method with  graph embedding methods, graph neural networks, and belief propagation methods.  

\begin{itemize}

\item \myparatight{Graph embedding methods} These methods first learn an embedding vector for each node; then they train a standard classifier (we consider multi-class logistic regression) using the training dataset and the embedding vectors; finally, they use the classifier to predict labels for the testing nodes. 
We consider two representative graph embedding methods, i.e.,  DeepWalk~\cite{perozzi2014deepwalk} and Planetoid~\cite{yang2016revisiting}.
We use 128-dimension embedding vectors.

\item \myparatight{Graph neural network methods} These methods learn the node embeddings and classify nodes simultaneously. In particular, we compare with Chebyshev~\cite{kipf2017semi}, graph convolutional network (GCN)~\cite{kipf2017semi}, GraphSAGE~\cite{hamilton2017inductive}, and graph attention network (GAT)~\cite{velivckovic2018graph}.

\item \myparatight{Belief propagation methods} We use the standard  BP~\cite{Pearl88} or the linearized BP (LinBP)~\cite{gatterbauer2015linearized} to perform inference in the pMRF. In particular, BP and LinBP infer the approximate marginal probability distributions $\mathbf{P}$ and use them to predict labels of testing nodes. We obtained the implementation of BP and LinBP from~\cite{gatterbauer2015linearized}.

\item \myparatight{Our method LCM and its variants}  In order to show the effectiveness of our consistency regularization term, we consider several variants of LCM.  Specifically, 
LCM-wo is the variant of our method that does not use the consistency regularization term. Moreover,   we compare our consistency regularization with the standard $L_1$ and $L_2$ regularizations, which are denoted as $\textrm{R}(\mathbf{W}, \mathbf{H})=\sum_{(u,v)\in E} |W_{uv}|_1 + \sum_{i,j} |H_{ij}|_1$ and $\textrm{R}(\mathbf{W}, \mathbf{H})=\sum_{(u,v)\in E} W_{uv}^2 + \sum_{i,j} H_{ij}^2$, respectively. 
Accordingly, we denote the two variants as LCM-L1 and LCM-L2, respectively. 

\end{itemize}

We observe that BP and LinBP outperform other traditional methods such as  label propagation (LP)~\cite{zhu2003semi}, manifold regularization (ManiReg)~\cite{belkin2006manifold}, deep semi-supervised embedding (SemiEmb)~\cite{weston2012deep}, and iterative classification algorithm (ICA)~\cite{lu2003link}. For instance, using the same experimental setting 
as Velickovic et al.~\cite{velivckovic2018graph}, LinBP achieves classification accuracies 0.785, 0.709, and 0.787 on the three citation graphs, respectively.  According to~\cite{velivckovic2018graph}, ICA consistently outperforms LP, ManiReg, and SemiEmb on the three citation graphs. However, ICA only achieves accuracies 0.751, 0.691, and 0.739 on the three citation graphs, respectively. Thus, we omit the results of these traditional methods for simplicity. 
The parameter settings of these methods are detailed in our supplementary material.

\begin{figure*}[!t]
\center
\subfloat[Cora]{\includegraphics[width=0.3\textwidth]{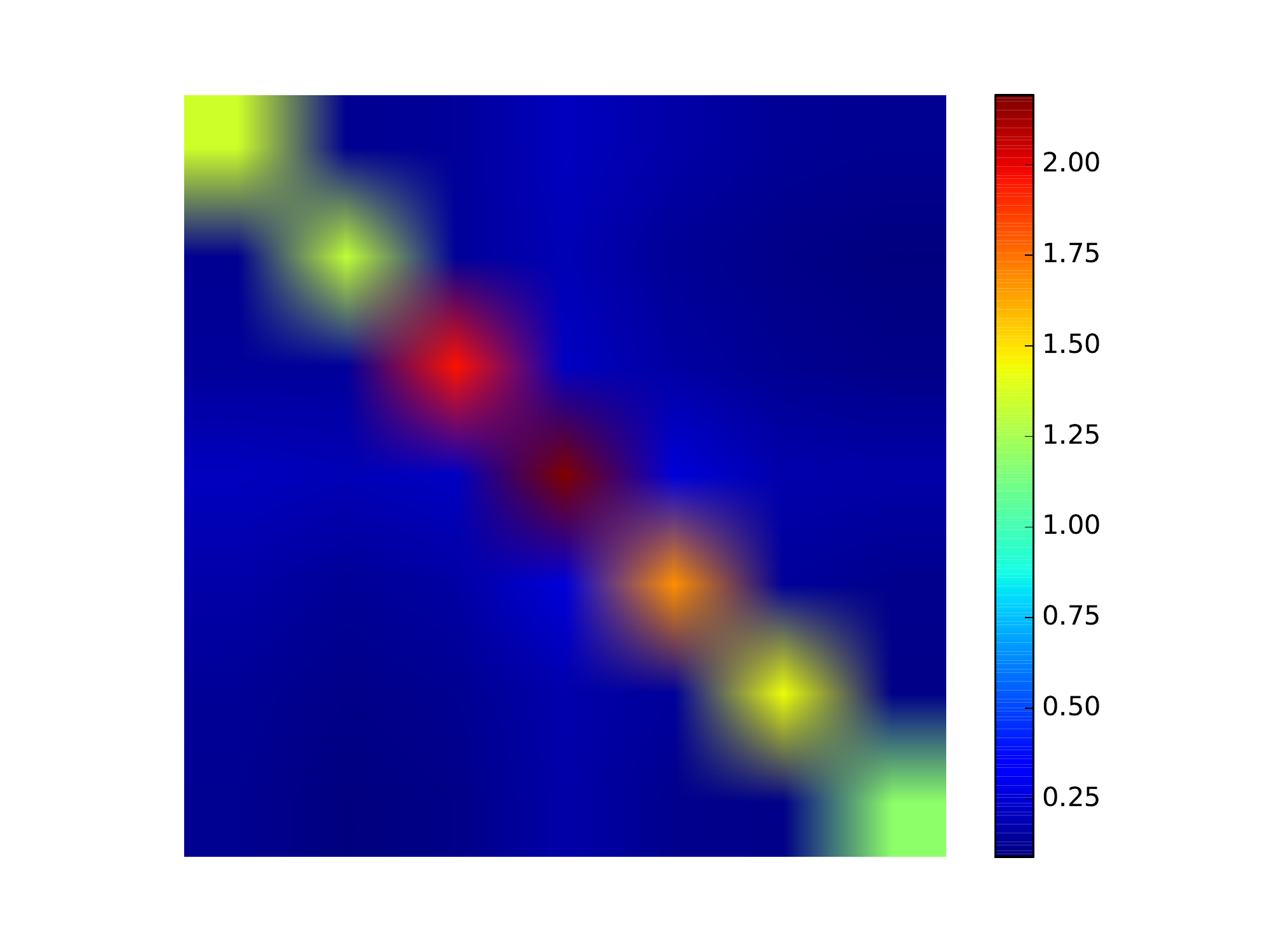}}
\subfloat[Citeseer]{\includegraphics[width=0.3\textwidth]{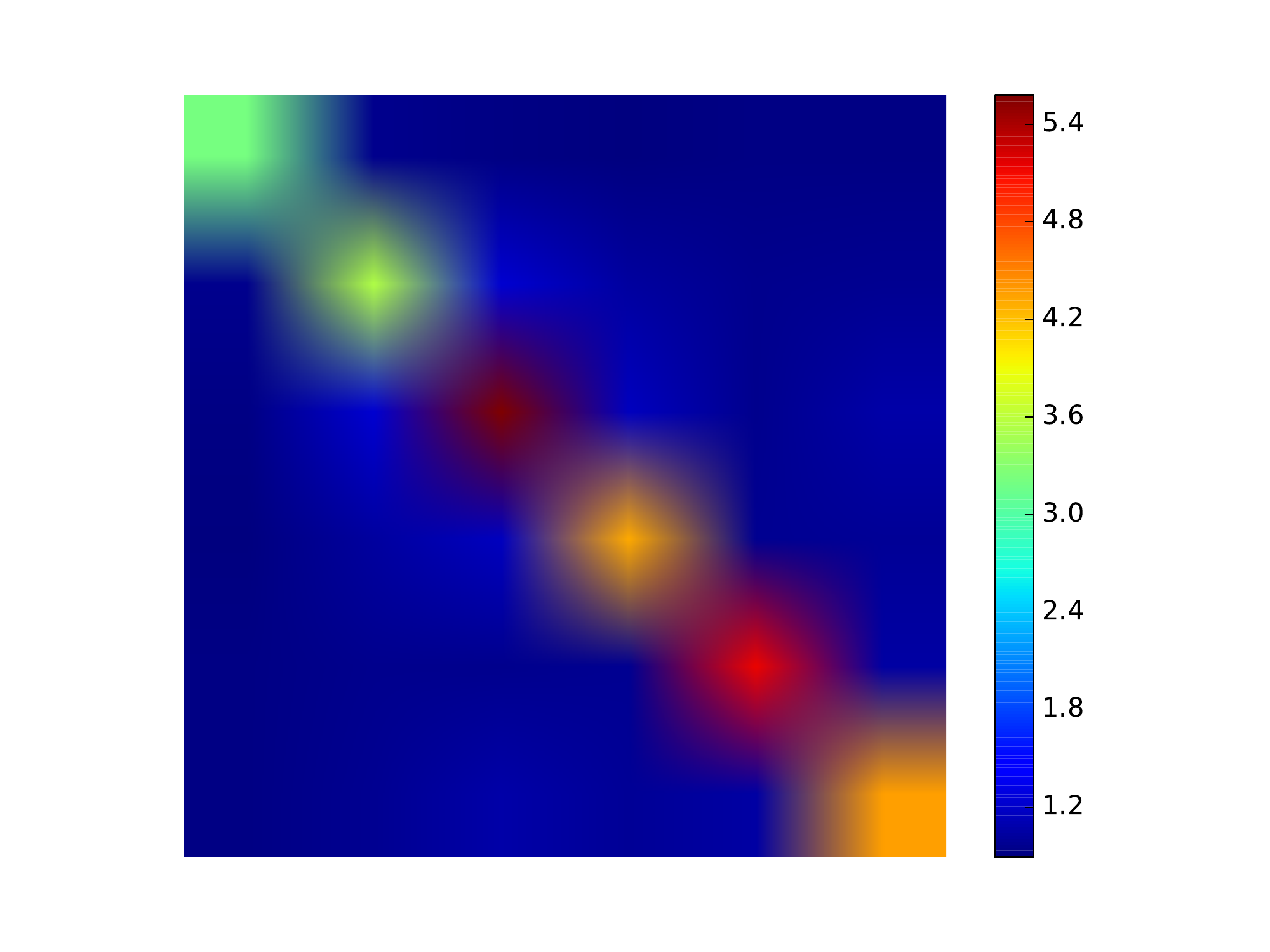}}
\subfloat[Pubmed]{\includegraphics[width=0.3\textwidth]{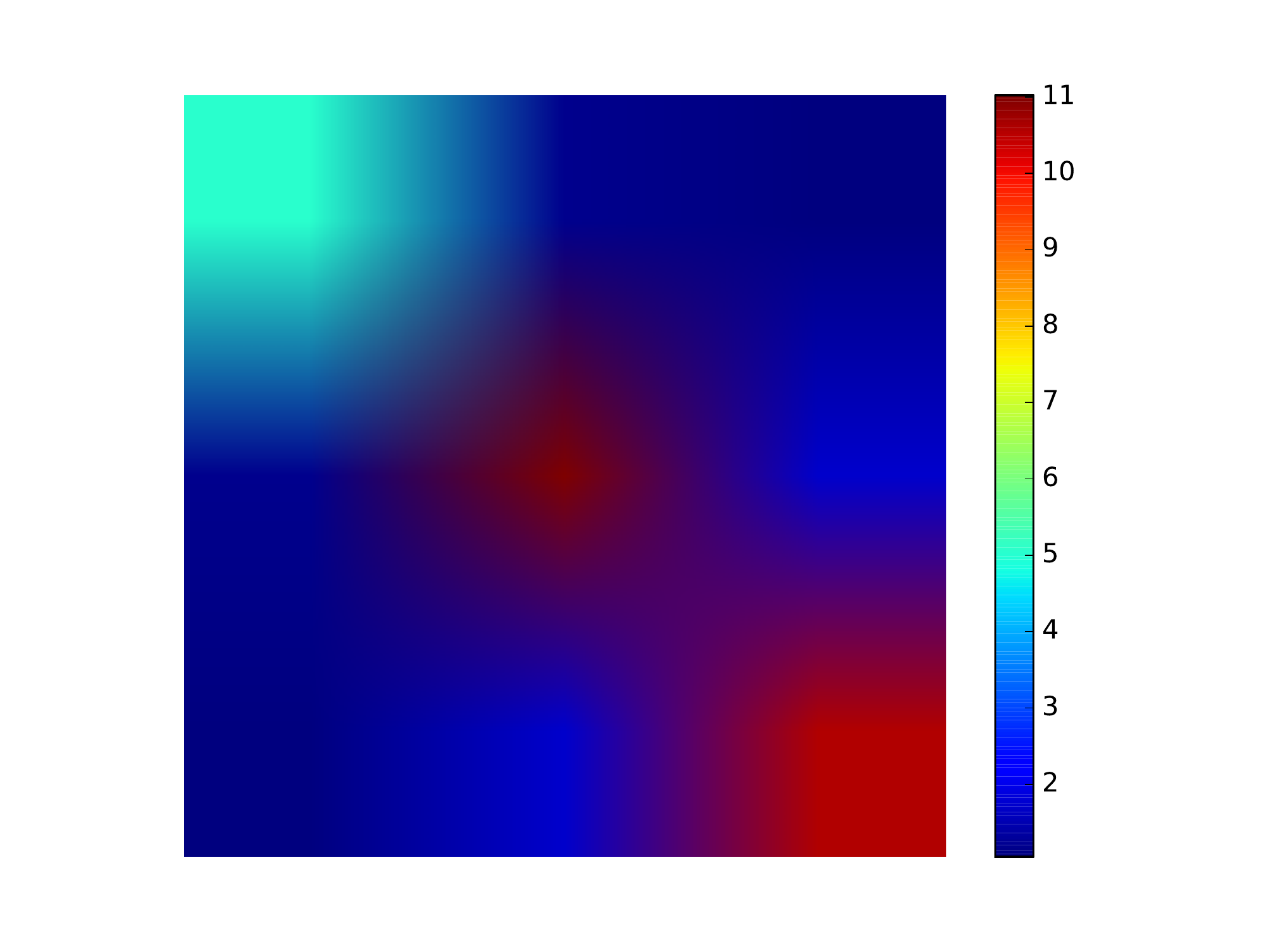}}
\\
\vspace{-4mm}
\subfloat[NELL]{\includegraphics[width=0.3\textwidth]{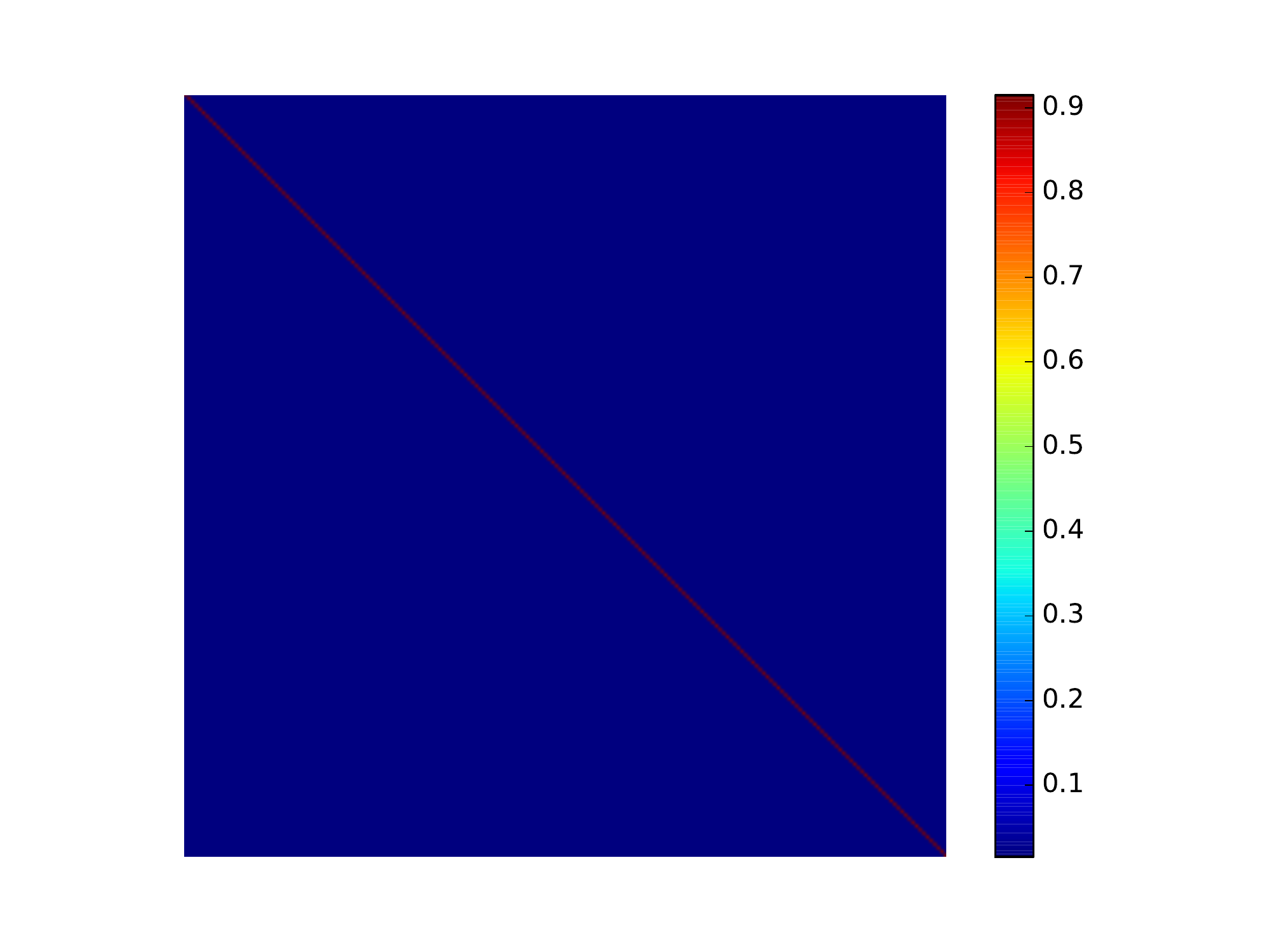}}
\subfloat[Google+]{\includegraphics[width=0.3\textwidth]{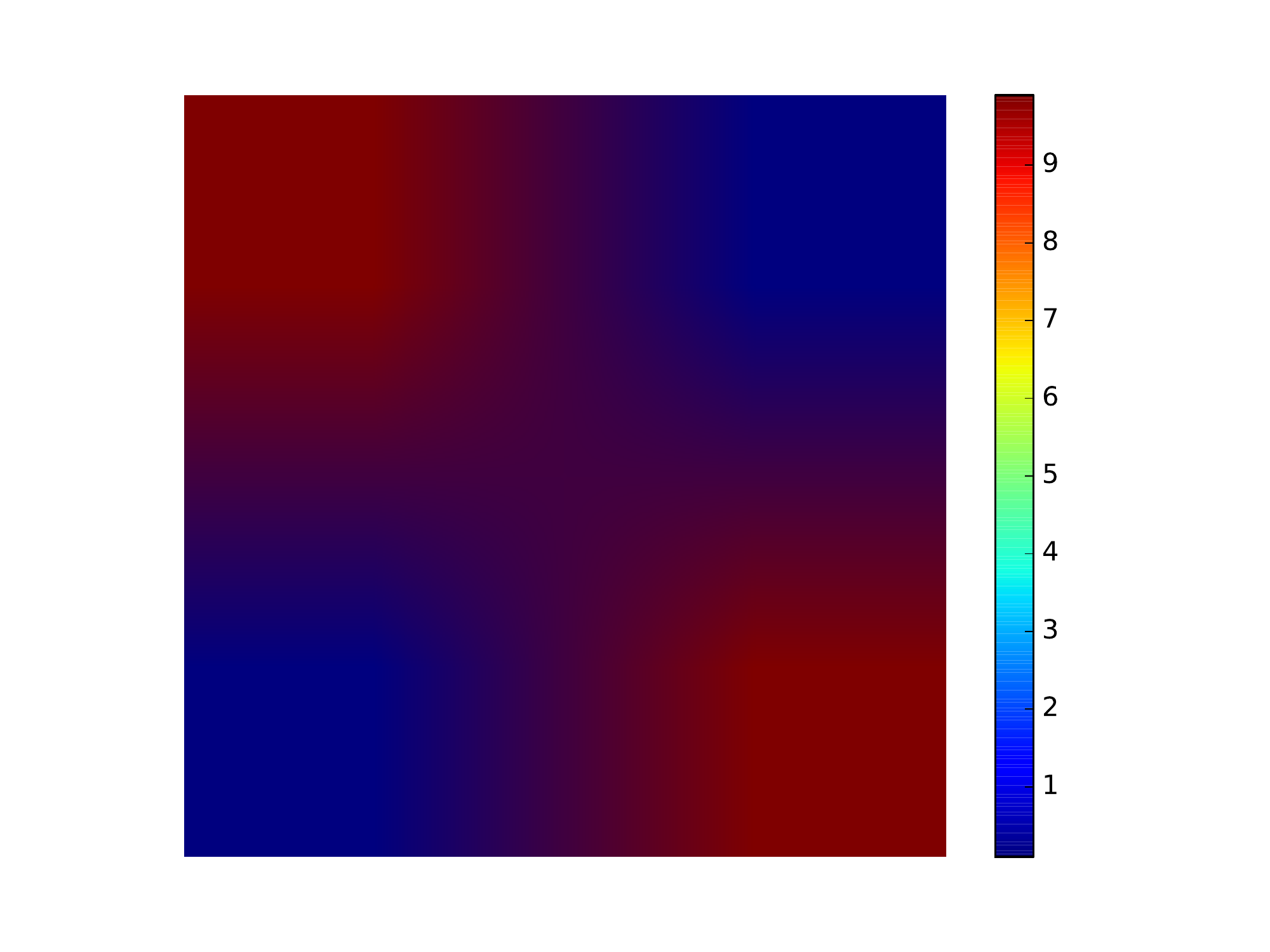}}
\subfloat[Twitter]{\includegraphics[width=0.3\textwidth]{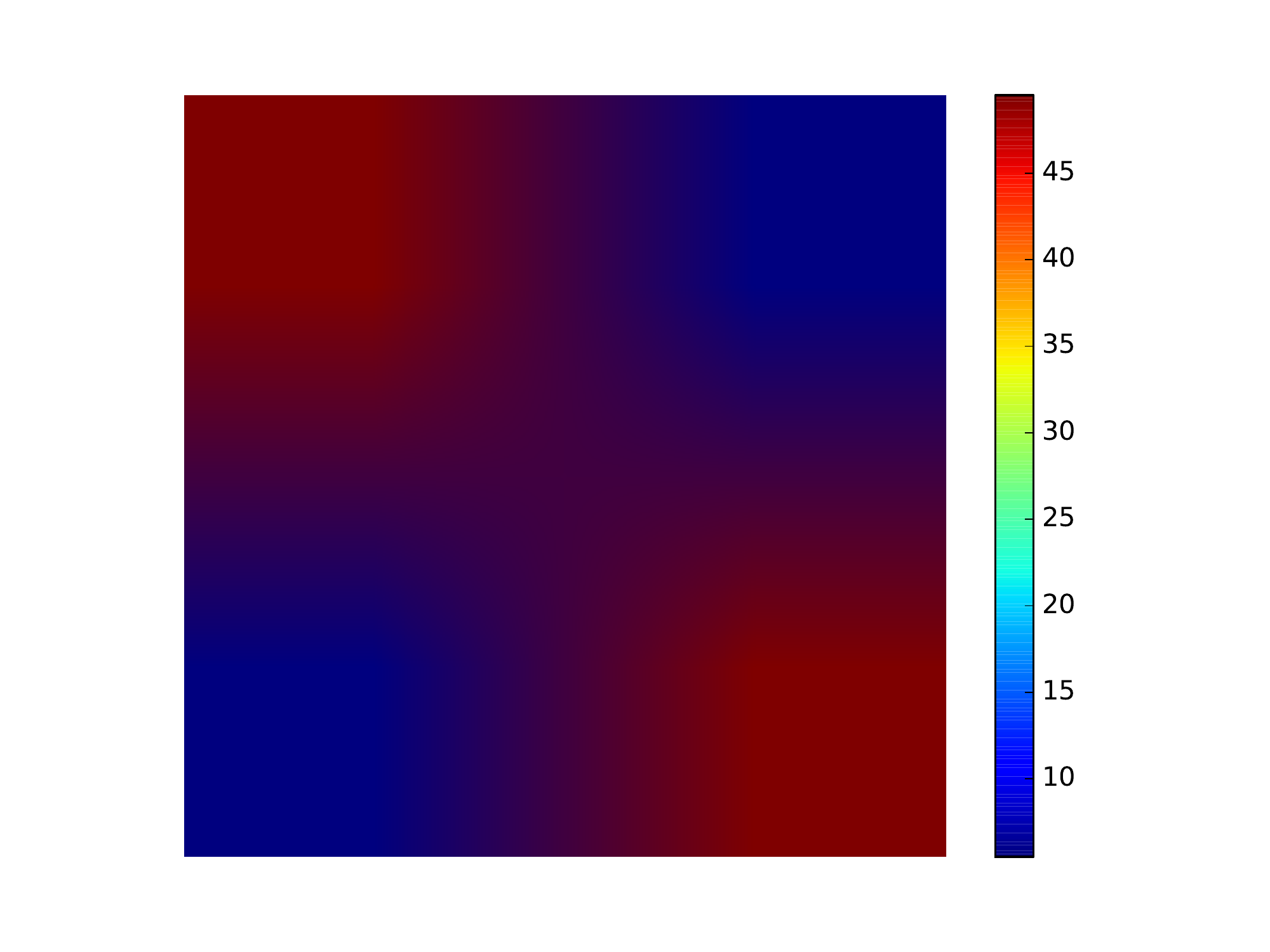}}
\caption{Learnt coupling matrices $\mathbf{H}$ on the six graphs. The coupling matrices are diagonal dominant. For the NELL graph, it is better to zoom in the subfigure to observe the diagonal dominant property of the learnt coupling matrix.}
\label{learnt_H}
\vspace{-4mm}
\end{figure*}

\begin{figure*}[!t]
\center
\subfloat[Cora]{\includegraphics[width=0.3\textwidth]{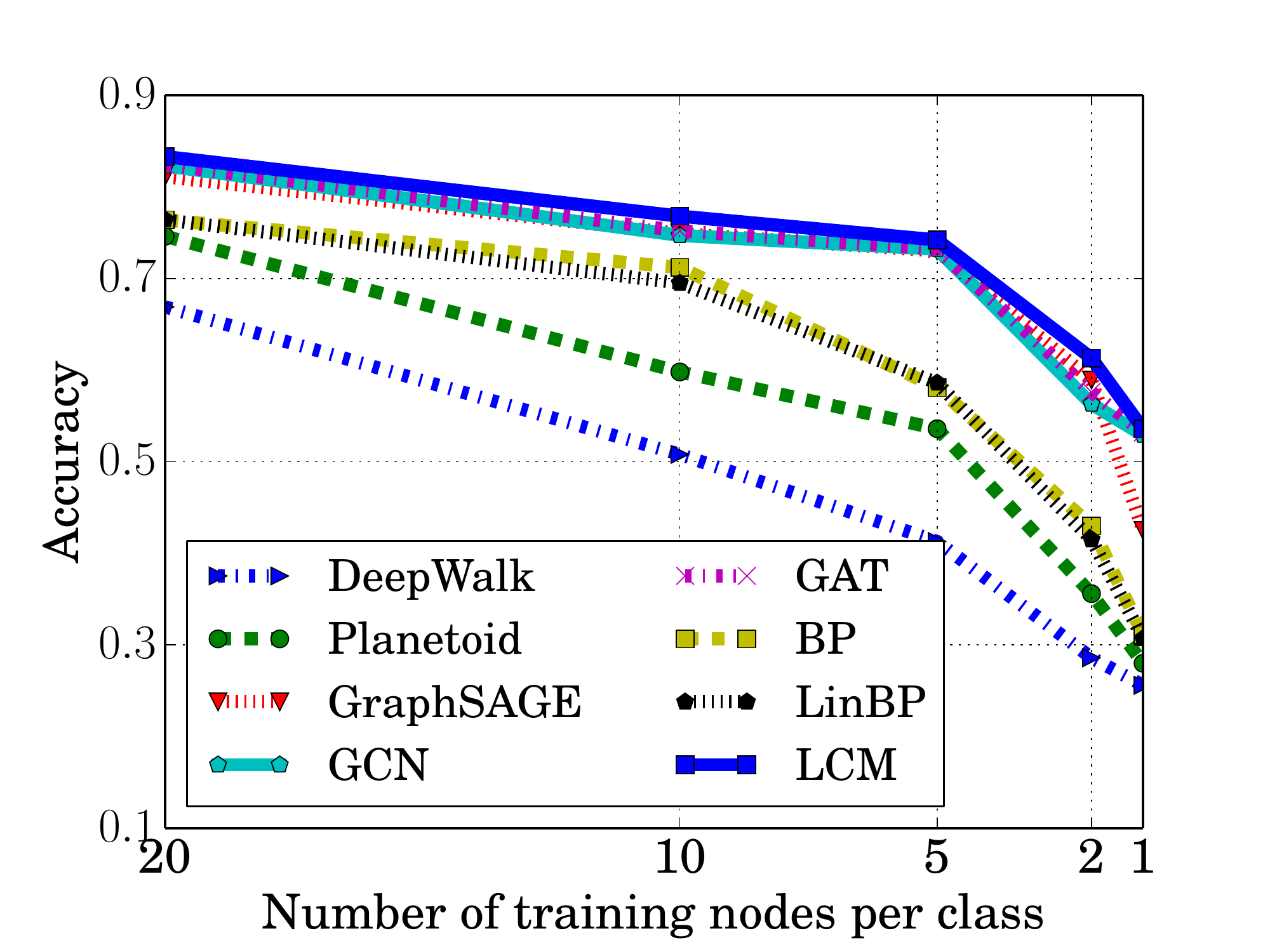}}
\subfloat[Citeseer]{\includegraphics[width=0.3\textwidth]{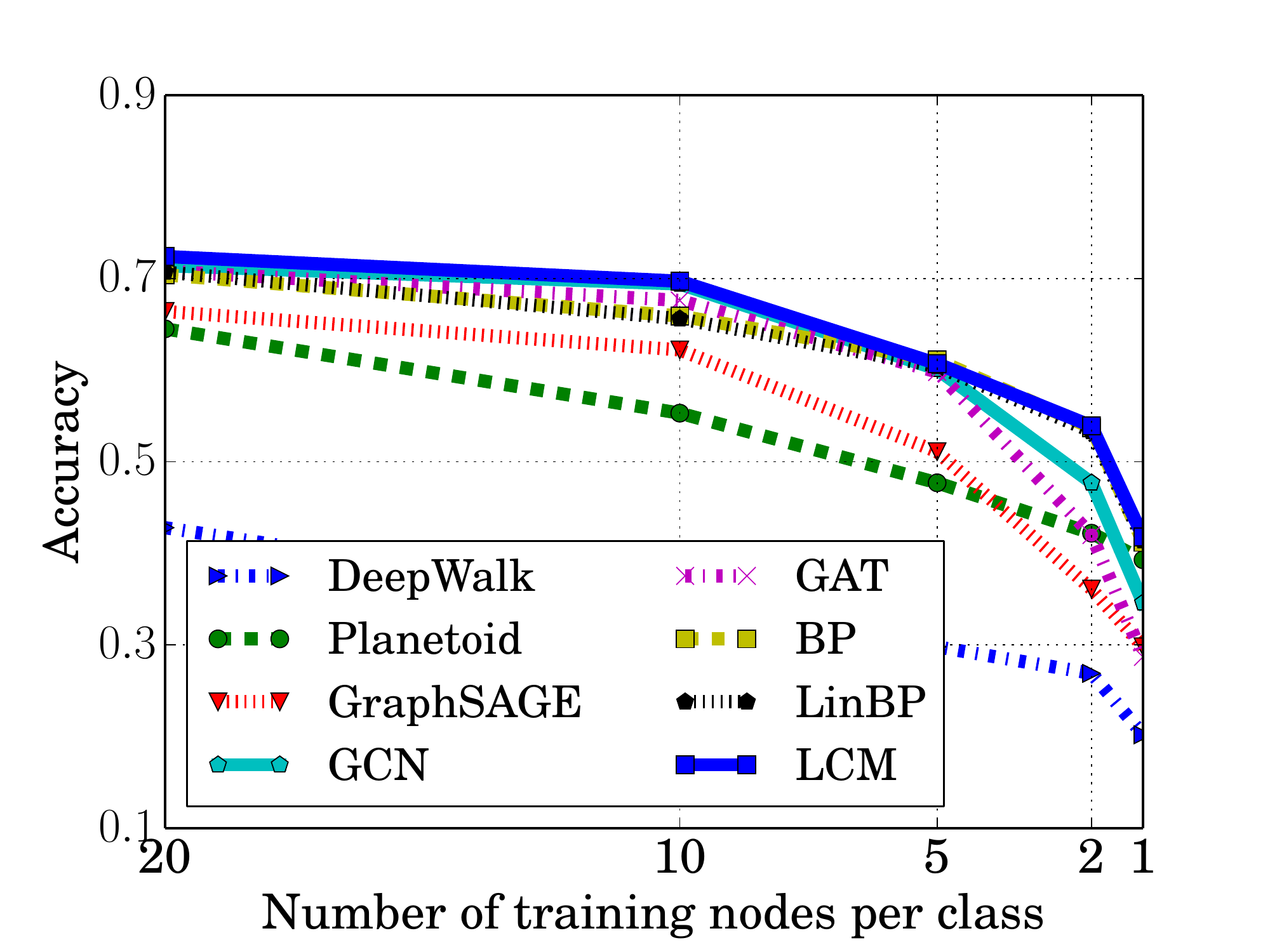}}
\subfloat[Pubmed]{\includegraphics[width=0.3\textwidth]{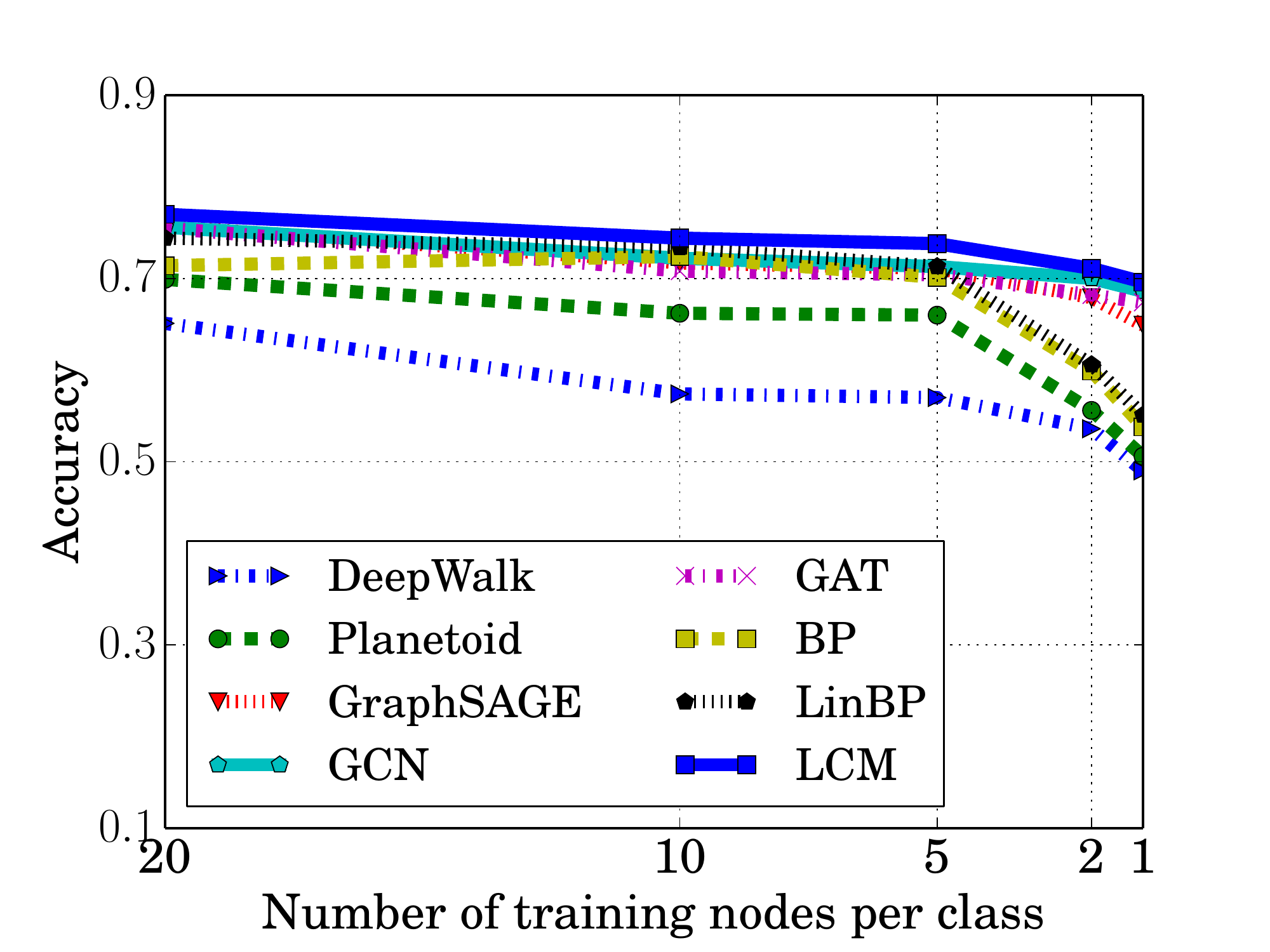}}
\\
\vspace{-4mm}
\subfloat[NELL]{\includegraphics[width=0.3\textwidth]{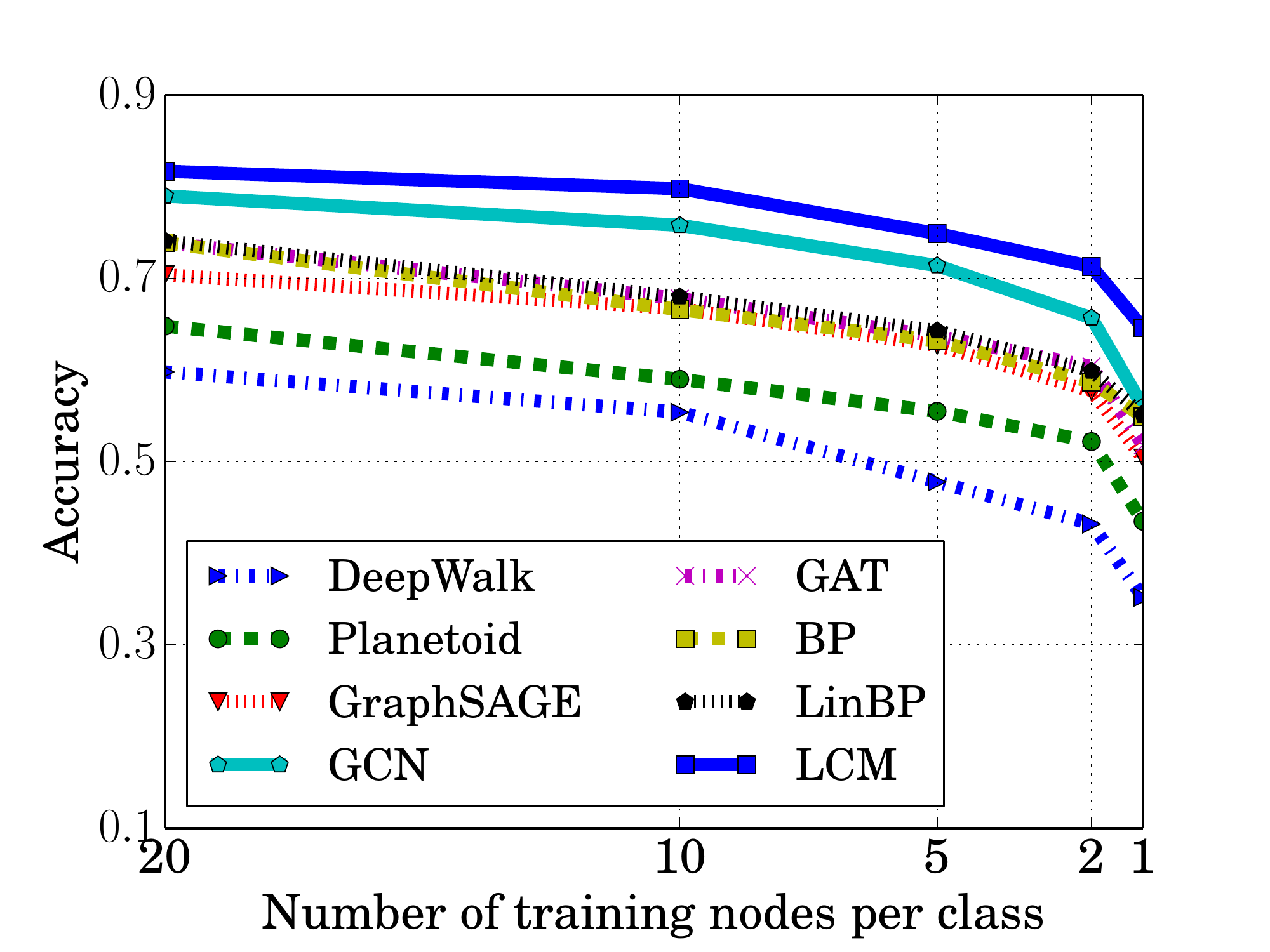}}
\subfloat[Google+]{\includegraphics[width=0.3\textwidth]{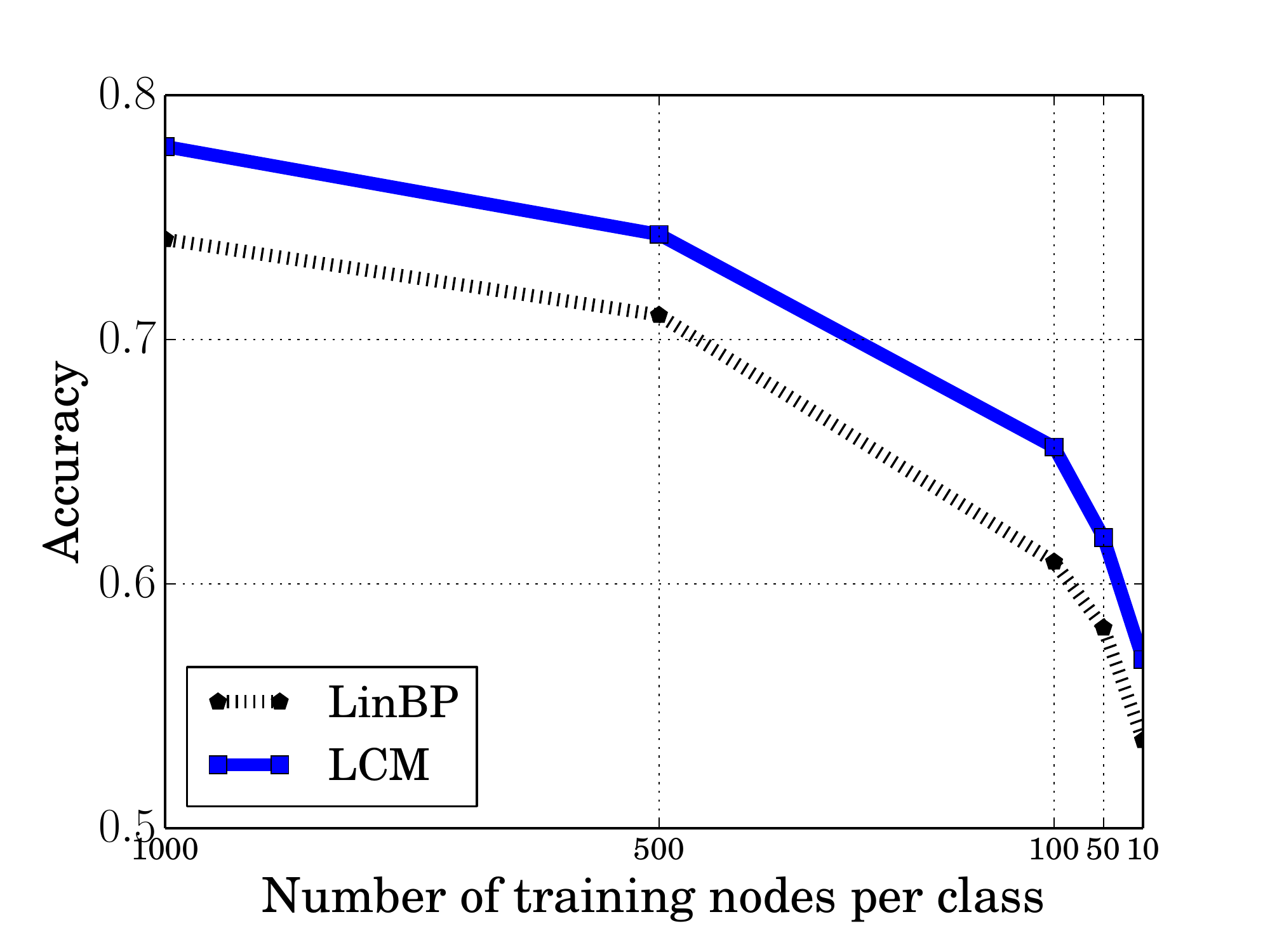}}
\subfloat[Twitter]{\includegraphics[width=0.3\textwidth]{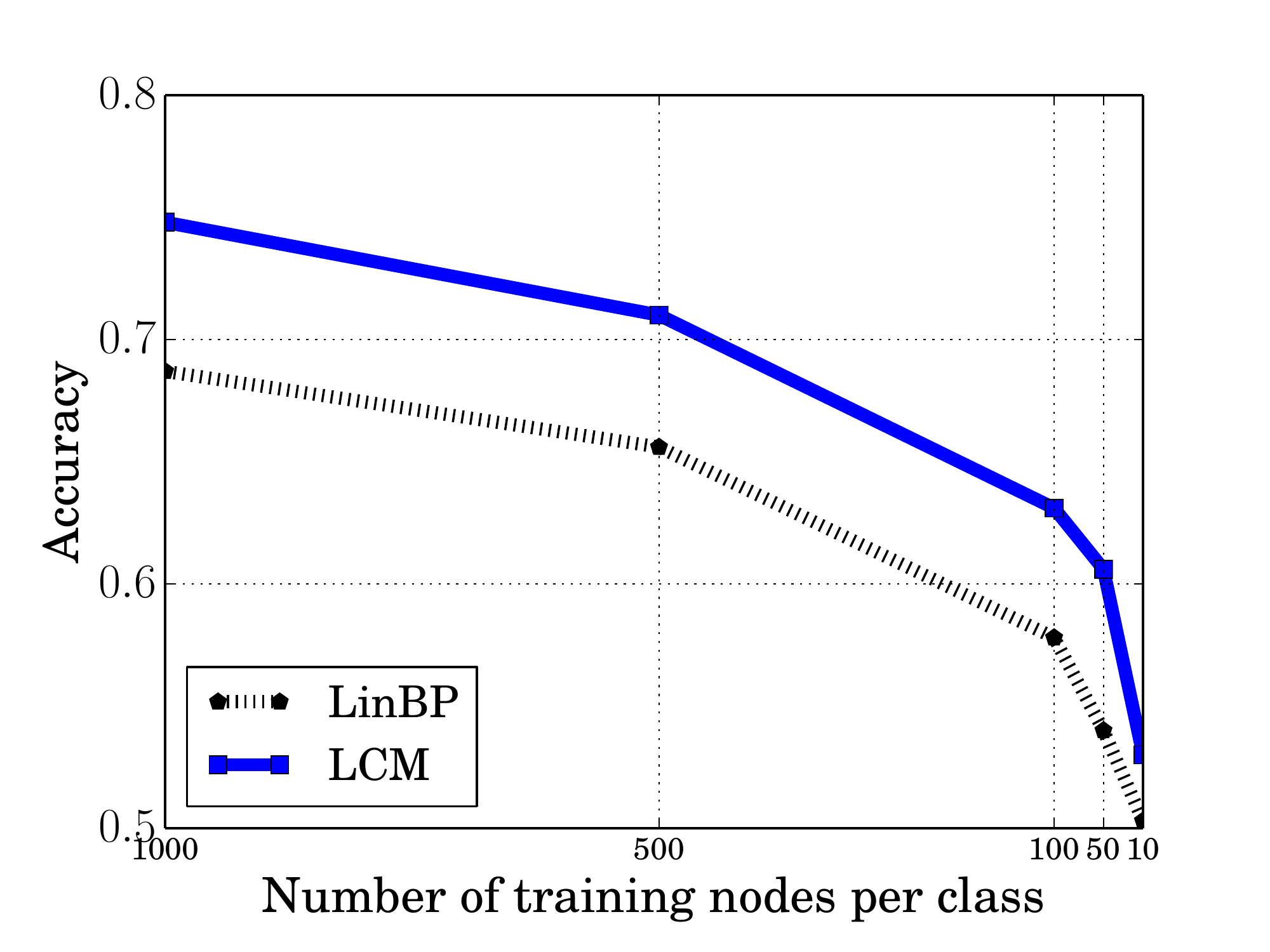}}
\caption{Impact of the number of training nodes per class. }
\label{impact_training}
\end{figure*}

\subsection{Results}

Table~\ref{performance} shows the classification accuracies averaged over 5 trials and the standard deviations of all compared methods on the six graphs. 
{We note that the compared graph embedding methods and graph neural networks cannot be executed on our machine for the two large-scale social graphs due to insufficient memory.
We have the following observations.}

\myparatight{LCM outperforms the compared methods} First, we observe that  graph neural networks outperform standard BP and LinBP methods. For example, GCN consistently achieves higher average accuracies than BP and LinBP on the three citation graphs and the NELL knowledge graph.  

Second, LCM is consistently more accurate than BP and LinBP. For example, LCM achieves 0.018 to 0.060 higher accuracies than BP on the three citation graphs and the NELL knowledge graph; and achieves 0.015 to 0.059 higher accuracies than LinBP on the six graphs.  
The reason is that BP and LinBP do not learn the edge weights and coupling matrix (i.e., they use the initialized edge weights and coupling matrix), while LCM does. To further illustrate the difference between LCM and BP/LinBP, Table~\ref{ave_weight} 
shows the average initialized weights and average weights learnt by LCM for homogeneous edges and heterogeneous edges on the six graphs.  An edge is homogeneous if the two corresponding nodes have the same label, otherwise the edge is heterogeneous. 
Moreover, Figure~\ref{learnt_H} shows the learnt coupling matrices for the six graphs.
Our learnt coupling matrices are diagonal dominant, which means that two linked nodes are more likely to have the same label.  Moreover, homogeneous edges have larger weights than heterogeneous edges on average after learning the edge weights. In particular, LCM increases edge weights for both  homogeneous edges and heterogeneous edges, but LCM increases the weights of homogeneous edges more substantially.

Third, after learning the weight matrix and the coupling matrix, our method outperforms the graph neural networks and graph embedding methods. Specifically, LCM achieves 0.009 to 0.051 higher average accuracies than the graph neural networks, and 0.071 to 0.212 higher average accuracies than the graph embedding methods on the three citation graphs and the NELL knowledge graph.

\myparatight{Our consistency regularization is effective} Our results also show that  our consistency regularization term is effective. In particular, LCM outperforms LCM-wo, LCM-L1, and LCM-L2. 
Our method achieves 0.006 to 0.036 higher accuracies when using the consistency regularization term (i.e., LCM vs. LCM-wo). 
{One possible reason is that L1/L2 regularization aims to prevent overfitting of complex models. However, pMRF is a simple model, which may not have overfitting on the datasets. Therefore, L1/L2 regularization could make LCM worse. Note that our consistency regularization does not aim to prevent traditional overfitting. Instead, it aims to improve the pMRF model itself and thus can outperform LCM-wo.  
}

\vspace{+2mm}
\myparatight{Impact of the number of training nodes per class} Figure~\ref{impact_training} further shows the average accuracies of different methods on the six graph datasets 
when we sample a particular number of training nodes from each class. Our method  consistently outperforms the compared methods when the training dataset size is small. 
{For instance, when each class has only 1 training node, LCM has an around 0.08 higher average accuracy than GCN on  NELL. When each class has 10 training nodes, LCM has an around 0.04 higher average accuracy than LinBP on Twitter.} 
Note that in semi-supervised node classification, we often assume the number of labeled nodes is small, compared to the total number of nodes.

\begin{table}[t]\renewcommand{\arraystretch}{1.0}
\centering
\begin{tabular}{|c|c|c|}
\hline
\multicolumn{2}{|c|}{\bf \textbf{Methods}} & {\bf \textbf{Time (seconds)}} \\ \hline
\multirow{2}{*}{\bf \makecell{Graph \\ Embedding}} 
& {\bf DeepWalk} & {1,312} \\  \cline{2-3}
& {\bf Planetoid} &  {1,033} \\  \hline \hline

\multirow{3}{*}{\bf \makecell{Graph \\ Neural \\ Network}} 
{\bf } & {\bf GCN} & {304}\\ \cline{2-3}
{\bf } & {\bf GraphSAGE} & {196} \\ \cline{2-3}
{\bf } & {\bf GAT} & {2,115}\\ \hline  \hline

\multirow{2}{*}{\bf \makecell{Belief \\ Propagation}} 
{\bf } & {\bf BP} & {443} \\ \cline{2-3}
{\bf } & {\bf LinBP}  & {8}\\ \hline\hline

\multirow{1}{*}{\bf \makecell{Our Method}} 
{\bf } & {\bf LCM} & {50} \\ \hline
\end{tabular} \\
\caption{Time of the compared methods on NELL.}
\label{efficiency}
\vspace{-4mm}
\end{table}

\myparatight{LCM is more efficient than graph embedding and graph neural network methods} 
For simplicity, we only show efficiency on NELL, which is the largest dataset {that the compared methods can be executed on our machine.}
Table~\ref{efficiency} shows the running time of the compared methods in one of our experiments. 
We have two observations. 
First, LCM is more efficient than graph embedding and graph neural network methods.  
Second, LCM is less efficient than LinBP. This is because LinBP does not learn the edge weight matrix and coupling matrix. However, both LCM and LinBP are much more efficient than BP. This is because 
BP needs to maintain messages on each edge.

%% file: related.tex
\section{Related Work}

\myparatight{Conventional methods} Semi-supervised node classification in graph-structured data has been studied extensively and many methods have been proposed. For instance, conventional methods include label propagation (LP)~\cite{zhu2003semi}, manifold regularization (ManiReg)~\cite{belkin2006manifold}, deep semi-supervised embedding (SemiEmb)~\cite{weston2012deep}, iterative classification algorithm (ICA)~\cite{lu2003link}, and pairwise Markov Random Fields (pMRF)~\cite{gatterbauer2017linearization}. In particular,  pMRF associates a discrete random variable with each node to model its label and defines a joint probability distribution for the random variables associated with all nodes, where the statistical correlations/independence between the random variables are captured by the graph structure. Then,  either the standard Belief Propagation (BP)~\cite{Pearl88} or linearized BP (LinBP)~\cite{gatterbauer2015linearized,gatterbauer2017linearization} is used for inference. Existing studies on pMRF-based semi-supervised node classification share a common limitation: they set a constant edge potential for all edges in the graph.   

\myparatight{Graph embedding methods and graph neural networks} 
A recent trend is to extend neural networks to graph-structured data for semi-supervised node classification. Methods along this direction can be roughly grouped into two categories, i.e., \emph{graph embedding}~\cite{perozzi2014deepwalk,tang2015line,cao2015grarep,yang2016revisiting,grover2016node2vec,ribeiro2017struc2vec,cui2020adaptive} and \emph{graph neural networks}~\cite{scarselli2009graph,duvenaud2015convolutional,atwood2016diffusion,kipf2017semi,hamilton2017inductive,velivckovic2018graph,xu2018powerful,battaglia2018relational,gao2018large,xu2018representation,ying2018hierarchical,qu2019gmnn,ma2019flexible,zhang2019heterogeneous,ma2019graph,wu2019net,chiang2019cluster,wang2020nodeaug}. 
Graph embedding methods first  learn node embeddings and then learn a standard classifier (e.g., logistic regression) using the embeddings to classify nodes, where learning the embeddings and learning the classifier are performed separately.  
Different graph embedding methods leverage different techniques to learn node embeddings. 
For instance, DeepWalk~\cite{perozzi2014deepwalk} learns nodes' embeddings via generalizing the word to vector technique~\cite{mikolov2013distributed}  developed
for natural language processing to graph data. 
Specifically, DeepWalk treats a node as a word in natural language, generates node sequences using truncated random walks on a graph, and leverages the skip-gram model~\cite{mikolov2013distributed} to learn an embedding vector for each node. 

Graph neural networks learn the node embeddings and the classifier to classify nodes simultaneously. In particular, the hidden layers represent node embeddings and the last layer models a classifier to classify nodes. Different graph neural networks use different neural network architectures. For instance, Graph Convolutional Network (GCN)~\cite{kipf2017semi} uses an architecture that is motivated by spectral graph convolutions~\cite{duvenaud2015convolutional}. Specifically, the input layer is the nodes' features. A hidden layer models nodes' embeddings. In particular,  the neural network iteratively computes a node's embedding vector in a hidden layer via aggregating the embedding vectors of the node's neighbors in the previous layer.  
State-of-the-art graph neural networks are more accurate than conventional methods such as LP and pMRF on multiple benchmark datasets.
However, graph neural networks are less efficient than conventional methods such as LP and LinBP. 
Our method addresses the limitation of pMRF-based methods and is more accurate and efficient than graph neural networks.

%% file: conclusion.tex
\section{Conclusion and Future Work}

We propose a novel method to learn the coupling matrix in pairwise Markov Random Fields for semi-supervised node classification in graph-structured data. We formulate learning coupling matrix as an optimization problem, whose objective function is the sum of the training loss and a consistency regularization term that we propose. Moreover, we propose an iterative algorithm to solve the optimization problem. Our evaluation results on six benchmark datasets show that our optimized pMRF-based method is more accurate and efficient than state-of-the-art graph neural networks. An interesting future work is to explore the connections between our optimized pMRF-based method and graph neural networks as well as unify them in a general framework.

\section{Acknowledgments}
We would like to thank the anonymous reviewers for their insightful reviews. This work was supported by the National Science
Foundation under grants No. 1937787 and 1937786. Any opinions,
findings and conclusions or recommendations expressed in this
material are those of the author(s) and do not necessarily reflect
the views of the funding agencies.

%% file: appendix.tex
\clearpage

\section{Dataset Description}
\myparatight{Citation graphs} 
In these graphs, nodes are documents and edges indicate citations between them, i.e.,  an edge between two documents is created if one document cites the other. The bag-of-words feature of a document is treated as the node feature vector. Each document also has a label. 

\myparatight{Knowledge graph} 
The  NELL dataset was extracted from the knowledge base introduced in~\cite{carlson2010toward}.  We use the version preprocessed by \cite{yang2016revisiting}. 
Specifically, nodes are entities or relations; an edge between an entity node and a relation node indicates the entity is involved in the relation with some other entity node. Like the citation graphs, the bag-of-words feature is treated as the node feature vector.  Each entity has a class label and there are 210 classes in total. 

\myparatight{Social graphs} 
Google+ is a social graph used to infer user's private attributes; and Twitter is a social graph used to perform fake account detection. Note that both Google+ and Twitter do not have node features. 
\begin{itemize}
\item {\bf Google+~\cite{jia2017attriinfer}:} This is an undirected social network with user attributes.
The dataset has around 5.7M users and 31M edges, where en edge between two users means that they are in each other's friend lists. Each user has an attribute \emph{cities lived} and the dataset contains 50 popular cities.
3.25\% of users disclosed at least one of these cities as their cities lived. Some users disclosed multiple cities. The task is to infer users' cities lived. 
Specifically, the dataset treats each city as a binary classification problem: given a city,  a user is treated as a \emph{positive} user if the user lives/lived in the city, and  a user is treated as a \emph{negative} user if he/she does not. 
Given a city, the task is to classify each user to be positive or negative depending on whether the user lives/lived in the city; and we perform the binary classification for the 50 cities separately.

\item {\bf Twitter:} 
This dataset was originally collected by \cite{kwak2010twitter}. 
Specifically, this dataset has around 41M nodes and 1.2B (undirected) edges, where a node is a Twitter user and an edge between two users means that they follow each other. 
The task is to classify an account/user to be fake or genuine. 
We obtained the ground truth node labels from \cite{wang2017sybilscar}.
Specifically, a fake account is an account/user that was suspended by Twitter, while a genuine account is an account/user that is still active on Twitter. 
In total, in the ground truth, 205,355 users are labeled as fake accounts, 36,156,909 users are labeled as genuine accounts, and the remaining users are unlabeled. 
\end{itemize}

\begin{table}[!t]\renewcommand{\arraystretch}{1.0}
\centering
\addtolength{\tabcolsep}{-3pt}
\begin{tabular}{|c|c|c|c|c|c|c|} \hline 
{\bf Dataset}  & {\bf \#Nodes} & {\bf \#Edges} & {\bf \#Features} & {\bf \#Classes} \\ \hline
{\bf Cora}  &  {\bf 2,708} & {\bf 5,429} & {\bf 1,433}   & {\bf 7} \\ \hline
{\bf Citeseer}  &  {\bf 3,327} & {\bf 4,732} & {\bf 3,703}   & {\bf 6} \\ \hline
{\bf Pubmed}  &  {\bf 19,717} & {\bf 44,338} & {\bf 500}   & {\bf 3}  \\ \hline
{\bf NELL}  & {\bf 65,755} & {\bf 266,144} &{\bf 5,414}   & {\bf 210}  \\ \hline
{\bf Google+} & {\bf 5,735,175} & {\bf 30,644,909} & {\bf N/A} & {\bf 2} \\ \hline
{\bf Twitter} &  {\bf 41,652,230} & {\bf 1,202,513,0464} & {\bf N/A}  & {\bf 2}  \\ \hline
\end{tabular}
\caption{Dataset statistics.}
\label{dataset_stat}
\end{table}

\section{Parameter Setting}
\label{para_set}

For our methods, we initialize the weight  matrix as ${W}_{uv}^{(0)}={1}/{\sqrt{d_ud_v}}$, where $d_u$ and $d_v$ are the degrees of nodes $u$ and $v$, respectively. 
Moreover, we initialize the coupling matrix  as $\tilde{H}_{i,i}^{(0)} = 0.9$ and $\tilde{H}_{i,j}^{(0)} = 0.1/(C-1)$ for $i \neq j$. This initialization means that two linked nodes are more likely to have the same label.   
{In the citation graphs and knowledge graph, we learn the probability distribution $\tilde{q}_v$ for each node  using a multi-class logistic regression classifier, the node features, and the training dataset, as we discussed in Section~\ref{backgroundpMRF}. 
In the social graphs, as nodes do not have features and node labels are binary, we assign the probability distribution $\tilde{q}_v$ for each node based on the training dataset alone. Specifically, we set $\tilde{q}_v = [1.0, 0.0]$ for a labeled positive node $v$ in the training dataset, $\tilde{q}_v = [0.0, 1.0]$ for a labeled negative node $v$ in the training dataset, and $\tilde{q}_v = [0.5, 0.5]$ for an unlabeled node $v$.
}
We perform 4 iterations of alternately updating $\mathbf{P}$ and learning $\mathbf{W}$ and $\mathbf{H}$. In each iteration of learning $\mathbf{W}$ and $\mathbf{H}$, we apply gradient descent 4 times. 
We use the validation dataset to tune the hyperparameters, i.e.,  $\gamma_1$, $\gamma_2$,  and $\lambda$. Specifically, we choose a set of values for each hyperparameter, i.e., $\gamma_1 \in \{0.02, 0.05, 0.1, 0.2\}$, $\gamma_2 \in \{0.0002, 0.0005, 0.001, 0.002\}$,  and $\lambda \in \{0.02, 0.05, 0.1, 0.2\}$. Then, we select the hyperparameters that achieve the highest accuracy on the validation dataset. We use smaller values for the learning rate $\gamma_2$ because the gradient $\frac{\partial \mathcal{L}(\mathbf{W}^{(t)}, {\mathbf{H}^{(t)}})}{\partial {H_{ij}^{(t)}}}$ is larger as it involves the sum over all edges. 
For other compared methods, we use their publicly available code and tune the associated hyperparameters. 
In particular, for DeepWalk, learning rate: \{0.1, 0.2, 0.4, 0.8\} and batch size: \{64, 128, 256, 512\}. For Planetoid, learning rate: \{0.01, 0.02, 0.05, 0.1\}, number of hidden units: \{25, 50, 75, 100\}, and batch size: \{50, 100, 150, 200\}. For Chebyshev, GCN, and GAT, learning rate: \{0.01, 0.02, 0.05, 0.1\}, number of hidden units: \{16, 32, 64, 128\}, and batch size: \{64, 128, 256, 512\}. For GraphSAGE, learning rate: \{0.0001, 0.001. 0.01, 0.1\} and batch size: \{64, 128, 256, 512\}. 
For BP and LinBP, number of iterations: \{5, 10, 15, 20\}, and 
we use the initialized weight  matrix and the coupling matrix without further learning them. 
All our experiments are performed on a Linux machine with 512GB memory and 32 cores.